\documentclass[10pt,journal,compsac]{IEEEtran}
\usepackage{amsmath,amsfonts}
\usepackage{array}
\usepackage[caption=false,font=normalsize,labelfont=sf,textfont=sf]{subfig}
\usepackage{textcomp}
\usepackage{stfloats}
\usepackage{url}
\usepackage{verbatim}
\usepackage{graphicx}
\usepackage{cite}
\usepackage{booktabs}
\usepackage{color}
\usepackage[linesnumbered,titlenumbered,ruled,vlined,commentsnumbered,noend]{algorithm2e}
\usepackage{comment}
\usepackage{multirow}
\usepackage{enumitem}
\usepackage{mathtools}
\usepackage{colortbl}
\usepackage{sidecap}
\usepackage{pifont}
\usepackage{flushend}
\usepackage{multicol}
\def\argmax{\operatornamewithlimits{arg\,max}}

\usepackage{xspace}
\makeatletter
\DeclareRobustCommand\onedot{\futurelet\@let@token\@onedot}
\def\@onedot{\ifx\@let@token.\else.\null\fi\xspace}
\def\eg{\emph{e.g}\onedot} 
\def\ie{\emph{i.e}\onedot} 
 
\def\etc{\emph{etc}\onedot} 
\def\wrt{w.r.t\onedot} 
\def\etal{\emph{et al}\onedot}
\makeatother

\definecolor{royalblue}{RGB}{65,105,225} %
\definecolor{lightblue}{RGB}{170,224,250} %
\definecolor{lightgreen}{RGB}{196,223,155}
\definecolor{lightyellow}{RGB}{254,247,153}
\usepackage[breaklinks=true,colorlinks,citecolor=royalblue,bookmarks=false]{hyperref}

\usepackage{scalerel}
\usepackage{tikz}
\usetikzlibrary{svg.path}
\definecolor{orcidlogocol}{HTML}{A6CE39}
\tikzset{
  orcidlogo/.pic={
    \fill[orcidlogocol] svg{M256,128c0,70.7-57.3,128-128,128C57.3,256,0,198.7,0,128C0,57.3,57.3,0,128,0C198.7,0,256,57.3,256,128z};
    \fill[white] svg{M86.3,186.2H70.9V79.1h15.4v48.4V186.2z}
                 svg{M108.9,79.1h41.6c39.6,0,57,28.3,57,53.6c0,27.5-21.5,53.6-56.8,53.6h-41.8V79.1z M124.3,172.4h24.5c34.9,0,42.9-26.5,42.9-39.7c0-21.5-13.7-39.7-43.7-39.7h-23.7V172.4z}
                 svg{M88.7,56.8c0,5.5-4.5,10.1-10.1,10.1c-5.6,0-10.1-4.6-10.1-10.1c0-5.6,4.5-10.1,10.1-10.1C84.2,46.7,88.7,51.3,88.7,56.8z};
  }
}
\newcommand\orcidicon[1]{\href{https://orcid.org/#1}{\mbox{\scalerel*{
\begin{tikzpicture}[yscale=-1,transform shape]
\pic{orcidlogo};
\end{tikzpicture}
}{|}}}}

\begin{document}

\title{Natural~\&~Adversarial Bokeh Rendering via Circle-of-Confusion Predictive Network}

\author{Yihao~Huang\,\orcidicon{0000-0002-5784-770X},
        Felix~Juefei-Xu\,\orcidicon{0000-0002-0857-8611},~\IEEEmembership{Member,~IEEE,}
        Qing~Guo$^\dagger$\,\orcidicon{0000-0003-0974-9299},~\IEEEmembership{Member,~IEEE,}
        Geguang~Pu\,\orcidicon{0000-0001-9750-8334}
        and~Yang~Liu\,\orcidicon{0000-0001-7300-9215},~\IEEEmembership{Senior~Member,~IEEE,}
\thanks{Yihao~Huang and Yang~Liu are with Nanyang Technological University, Singapore. Felix~Juefei-Xu is with New York University, USA. Qing~Guo is with the Institute of High Performance Computing (IHPC) and Centre for Frontier AI Research (CFAR), Agency for Science, Technology and Research (A*STAR), Singapore. Geguang~Pu is with 1) East China Normal University and 2) Shanghai Industrial Control Safety Innovation Technology Co., LTD, China. $\dagger$ Qing Guo is the corresponding author (tsingqguo@ieee.org).}
}




\maketitle

\begin{figure*}[tb]
\centering
\includegraphics[width=\linewidth]{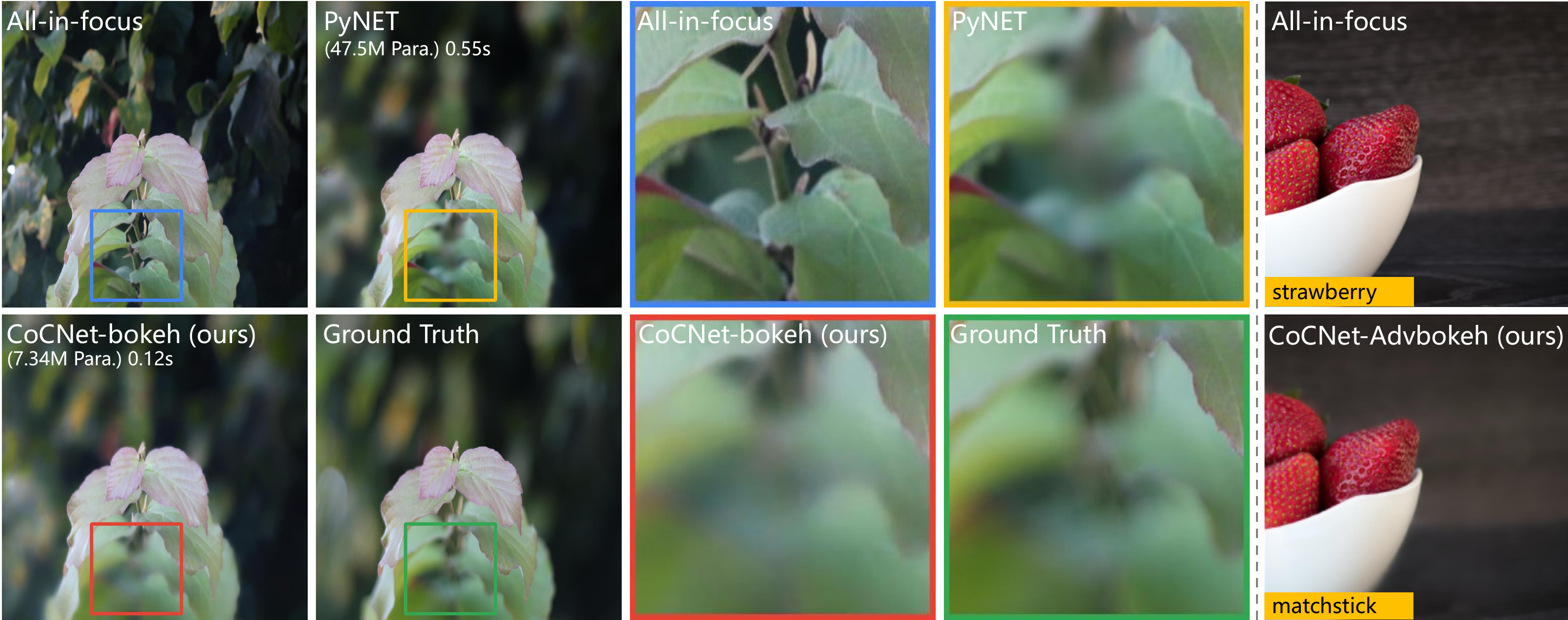}
\caption{In the left part shows the comparison between CoCNet (ours) and state-of-the-art method PyNET \cite{ignatov2020rendering}. CoCNet has a smaller model size and needs less inference time based on a single Tesla V100 GPU. The right part shows the adversarial bokeh example generated by CoCNet. CoCNet-AdvBokeh successfully fools the ResNet50 model with the adversarial bokeh effect.}
\label{fig:teaser}
\end{figure*}

\begin{abstract}
Bokeh effect is a natural shallow depth-of-field phenomenon that blurs the out-of-focus part in photography.
In recent years, a series of works have proposed automatic and realistic bokeh rendering methods for artistic and aesthetic purposes. They usually employ cutting-edge data-driven deep generative networks with complex training strategies and network architectures.
However, these works neglect that the bokeh effect, as a real phenomenon, can inevitably affect the subsequent visual intelligent tasks like recognition, and their data-driven nature prevents them from studying the influence of bokeh-related physical parameters (\ie, depth-of-the-field) on the intelligent tasks.
To fill this gap, we study a totally new problem, \ie, \textit{natural~\&~adversarial bokeh rendering}, which consists of two objectives: rendering realistic and natural bokeh and fooling the visual perception models (\ie, bokeh-based adversarial attack).
To this end, beyond the pure data-driven solution, we propose a hybrid alternative by taking the respective advantages of data-driven and physical-aware methods.
Specifically, we propose the \textit{circle-of-confusion predictive network (CoCNet)} by taking the all-in-focus image and depth image as inputs to estimate circle-of-confusion parameters for each pixel, which are employed to render the final image through a well-known physical model of bokeh.
With the hybrid solution, our method could achieve more realistic rendering results with the naive training strategy and a much lighter network. 
Moreover, we propose the adversarial bokeh attack by fixing the CoCNet while optimizing the depth map \wrt the visual perception tasks. Then, we are able to study the vulnerability of deep neural networks according to the depth variations in the real world. 
The extensive experiments show that our method produces more realistic bokeh than the state-of-the-art methods while fooling the powerful deep neural networks with a high accuracy drop. 
\end{abstract}

\begin{IEEEkeywords}
Bokeh Rendering, Circle-of-Confusion, Adversarial Attack
\end{IEEEkeywords}

\IEEEpeerreviewmaketitle


\section{Introduction}\label{sec:intro}
In photography, the shallow depth-of-field (DoF) effect or the bokeh effect is an important technique to generate aesthetically pleasing photographs. The images with the bokeh effect draw the attention of the viewers by primarily blurring the out-of-focus parts of the images while keeping the focused parts sharp. This effect can wash out unnecessary image details such as cluttered backgrounds, which can save the viewers from the messy information of the images and emphasize the salient themes such as the foreground person or object of the images.
There are mainly two ways to produce a bokeh effect, which is either generated optically or through computational photography methods. To physically and optically produce the bokeh effect, usually fast lenses with large apertures in tandem with high-end digital single-lens reflex (DSLR) cameras or the latest digital single-lens mirrorless (DSLM) cameras are needed. The high barrier of entry has limited this option largely to professionals. To benefit more people, the latest smartphone manufacturers have tried to enable a computational photography way of generating a realistic bokeh effect on consumer-level cell phone products, which strongly drives research in this direction.

In recent years, a lot of works \cite{xiao2018deepfocus,wang2018deeplens,dutta2021depth,ignatov2020rendering,busam2019sterefo,wadhwa2018synthetic,dutta2021stacked,qian2020bggan,luo2020bokeh} have been proposed and ``AIM 2019 Challenge on Bokeh Effect Synthesis'' competitions \cite{ignatov2019aim} have been organized to promote the development of computational photography-based bokeh effect synthesis. These methods usually employ data-driven deep generative networks with complex training strategies and network architectures. The state-of-the-art methods (\eg, PyNET \cite{ignatov2020rendering} and DMSHN \cite{dutta2021stacked}) are typical examples.

However, as a kind of common image style, the bokeh effect inevitably influences the subsequent visual tasks. To further study the relationship between the physical parameters of the bokeh effect with the visual intelligence tasks, we raise a new problem (\ie, natural \& adversarial bokeh rendering) which aims to generate natural and realistic bokeh effect and fool the visual perception model. It is obvious that pure data-driven methods can hardly support such research due to their physical-independent property. Thus we propose a hybrid bokeh synthesis network that is guided by the advantages of data and physical principles.


We first analyze the bokeh generation process of the camera and summarize the physical model of the bokeh. Then, by imitating the operation form as priori, we propose a novel \textbf{\emph{circle-of-confusion predictive network (CoCNet)}} that takes the all-in-focus image and depth image as inputs to estimate circle-of-confusion (CoC) parameters. The method is able to realistically synthesize the bokeh effect, with a simple training strategy and lightweight design. Specifically, the way of synthesis is implemented by fusing the all-in-focus image with learned templates. The templates are generated by filtering the all-in-focus image with specialized kernels predicted in the network, which simulates the physical rendering process by implicitly predicting the CoC for each pixel. Through experiments, we verify that our proposed method can adaptively estimate the kernels according to the image content and achieve comparable performance to the state-of-the-art methods with fewer model parameters (\ie, less than $\frac{3}{4}$). In the case of simplifying the convolution channels of the model, CoCNet can achieve on-par performance to the state-of-the-art methods with fewer than merely $\frac{1}{5}$ parameters.
%


Due to the natural advantage and universality of bokeh in photos, as well as potential adverse effects on multifarious visual tasks, in this work, we also aim to reveal the vulnerabilities in various visual image understanding tasks with the help of CoCNet, with a newly proposed task termed \textbf{\emph{adversarial bokeh attack (AdvBokeh)}}. The method embeds deceptive information into the bokeh generation procedure and produces a natural bokeh-like adversarial example without any human-noticeable noise artifacts by fixing the parameters of CoCNet while optimizing the depth map. Adversarial bokeh examples are shown in Fig.~\ref{fig:teaser} on the rightmost panel.
We validate the proposed method on a popular adversarial image classification dataset, \ie, NeurIPS-2017 DEV, and show that the proposed method can penetrate four state-of-the-art (SOTA) image classification networks, \ie, ResNet50, VGG, DenseNet, and MobileNetV2 with high success rates as well as high image quality. Moreover, the adversarially generated defocus blur images from the AdvBokeh can actually be exploited to enhance the performance of downstream tasks (\eg, defocus deblurring system), which shows the versatility of the adversarial bokeh samples.

The contributions are summarized as follows:

\ding{182} We raise a new computer vision problem (\ie, natural \& adversarial bokeh rendering) for investigation and propose {\emph{circle-of-confusion predictive network (CoCNet)}} by combining the advantages of data-driven and physical-aware methods. It achieves on-par SOTA performance with a more lightweight design and a much simpler training strategy.

\ding{183} We propose the depth-guided attack, \ie, AdvBokeh for image classification tasks by tapping into the bokeh generation process via the proposed CoCNet. The generated adversarial bokeh images can be used to improve the performance of the SOTA defocus deblurring systems, \ie, IFAN \cite{Lee_2021_CVPR}, demonstrating that the adversarial bokeh examples can further enhance downstream visual tasks.  

\ding{184} We compare the performance of CoCNet with the state-of-the-art methods (\ie, PyNET, DMSHN, MPFNet, BRViT) on the popular EBB! bokeh generation dataset. The adversarial attack experiments are carried out on a popular adversarial image classification dataset (\ie, NeurIPS-2017 DEV), showing that the proposed method can penetrate four classical image classification networks with high success rates as well as maintain high image quality. The adversarial examples obtained by AdvBokeh also exhibit a certain degree of transferability under black-box settings. 

We introduce related works in Sec.~\ref{sec:related}. In Sec.~\ref{sec:bokeh_synthesis_method}, we use a lot of figures and formulas to illustrate the design of CoCNet and the implementation architecture. In Sec.~\ref{sec:influence_on_visual_task}, we propose how to generate a natural~\&~adversarial bokeh effect with CoCNet. In Sec.~\ref{sec:exp}, we conduct quantitative and qualitative experiments to verify the effectiveness of our method. The conclusion of the paper is in Sec.~\ref{sec:concl}.

\noindent\textbf{Scope.}
Our research belongs to the image synthesis task in the multimedia domain. We raise a new image synthesis problem (\ie, natural \& adversarial bokeh rendering) for investigation. The research can help to reveal the vulnerabilities in various visual multimedia understanding tasks under bokeh rendering.


\section{Related Work}\label{sec:related}

\subsection{DoF Rendering}
DoF rendering plays an important role in realistic image synthesis \cite{xu2021virtual,zhang2021pr,meng2023automatic,wan2022benchmarking}. A number of works \cite{haeberli1990accumulation,lee2010real,soler2009fourier,wu2010realistic,yu2010real} have tried using ray tracing or light field rendering to synthesize realistic bokeh effect images. Most of these methods need accurate 3D scene information and are time-consuming. 
%

%
To achieve a realistic bokeh effect, early works \cite{shen2016deep,shen2016automatic,xu2018rendering} first take a portrait as the target of bokeh effect rendering. 
%
%
However, the previous methods put emphasis on portrait photos, which neglects the bokeh effect generation for images in the wild.
To expand the application domain of DoF rendering, recently researchers proposed a lot of works \cite{xiao2018deepfocus,wang2018deeplens,dutta2021depth,ignatov2020rendering,busam2019sterefo,wadhwa2018synthetic,dutta2021stacked,qian2020bggan,luo2020bokeh,purohit2019depth}. Some of them \cite{xiao2018deepfocus, wang2018deeplens, busam2019sterefo, wadhwa2018synthetic, purohit2019depth} are starved of fine depth maps collected by the additional sensors or calculated by pretrained depth estimation network. Wang \etal{} \cite{wang2018deeplens} propose a neural network with a depth estimation module, a lens blur module, and a guided upsampling module to render the bokeh effect on high-resolution images. 
%

Since depth estimation is runtime-intensive and not suitable for consumer-level phones. Thus the ``AIM 2019 Challenge on Bokeh Effect Synthesis'' competition \cite{ignatov2019aim} asks the participants to generate bokeh images with only one single frame. The recent works \cite{dutta2021depth,ignatov2020rendering,dutta2021stacked,qian2020bggan,luo2020bokeh,wang2022self,seizinger2023efficient,yang2023bokehornot,jeong2020real} have tried to directly render the bokeh effect by a well-designed network without the help of the depth map. 
Dutta \etal{} \cite{dutta2021depth} propose to fuse the all-in-focus image with a Gaussian-blurred version of it. However, Gaussian blur is physically different from the blur effect of bokeh and preparing the Gaussian-blurred version of an all-in-focus image is time-consuming. PyNET \cite{ignatov2020rendering}, DMSHN \cite{dutta2021stacked}, MPFNet \cite{wang2022self}, and BRViT \cite{nagasubramaniam2022bokeh} achieve the state-of-the-art performance.
Ignatov \etal{} \cite{ignatov2020rendering} present a large-scale bokeh dataset named ``Everything is Better with Bokeh!'' (EBB!), which contains 5K shallow DoF image pairs captured using the Canon 7D DSLR. They propose PyNET-based architecture with multi-stage training to render the bokeh effect. Dutta \etal{} \cite{dutta2021stacked} proposes an end-to-end deep multi-scale hierarchical network (DMSHN) for the bokeh effect rendering of images captured from the monocular camera, also with multi-stage training. BRViT \cite{nagasubramaniam2022bokeh} uses an end-to-end pyramid ViT as the backbone for Bokeh rendering of images from a monocular camera. MPFNet \cite{wang2022self} uses a self-supervised multi-scale pyramid fusion network to generate bokeh images. The pyramid architecture used by BRViT and MPFNet is complex and unwieldy. In summary, these bokeh rendering methods design complicated networks and need a time-consuming multi-stage training strategy to render the bokeh effect, which is not physically aware and efficient.

\subsection{Unrestricted Attack}
Adversarial attacks \cite{zhang2022transferable,wan2023average} that generate $L_{p}$-norm perturbations have obvious noises and are now considered unrealistic. Recent works \cite{shamsabadi2020colorfool,engstrom2019exploring,guo2020watch,cheng2021pasadena} have tried to put emphasis on generating unrestricted adversarial images, which will not raise the suspicion of the people. From this point of view, unrestricted adversarial images make more sense in practice.
The attacks mainly focus on three categories: geometric transformation, color modification and photography effect. For geometric transformation, a few works exploit image deformations to construct adversarial attacks \cite{engstrom2019exploring}. The images do not contain unnatural noise. However, image distortion is not natural when applied to images with straight edges. For color modification, some works have tried to do semantic adversarial attacks by modifying the color of the object in the image \cite{shamsabadi2020colorfool, zhao2020adversarial}. However, the colors of some objects in the adversarial attacked images defy common sense and look very fake (\eg, yellow river, purple tree, \etc). 
For photography effect, Guo \etal \cite{guo2020watch} propose an adversarial attack method that can generate visually motion-blurred adversarial examples, which mimics a type of photographic effect with high fidelity. However, the method cannot be extended to generate defocus blur.


\section{Bokeh Synthesis Method}\label{sec:bokeh_synthesis_method}

In Sec.~\ref{sec:mot}, we introduce the motivation for designing a hybrid bokeh rendering network. To achieve this, in Sec.~\ref{sec:camera_bokeh_modeling}, we infer the realistic bokeh imaging model based on the optical principle of a normal sensor. However, rendering that strictly adheres to the physical model is time-consuming, thus in Sec.~\ref{sec:Circle-of-confusion_Predictive_Network}, we propose two time-efficient approximation strategies and design a lightweight bokeh synthesis network.

\begin{figure}
\centering
\includegraphics[width=0.8\linewidth]{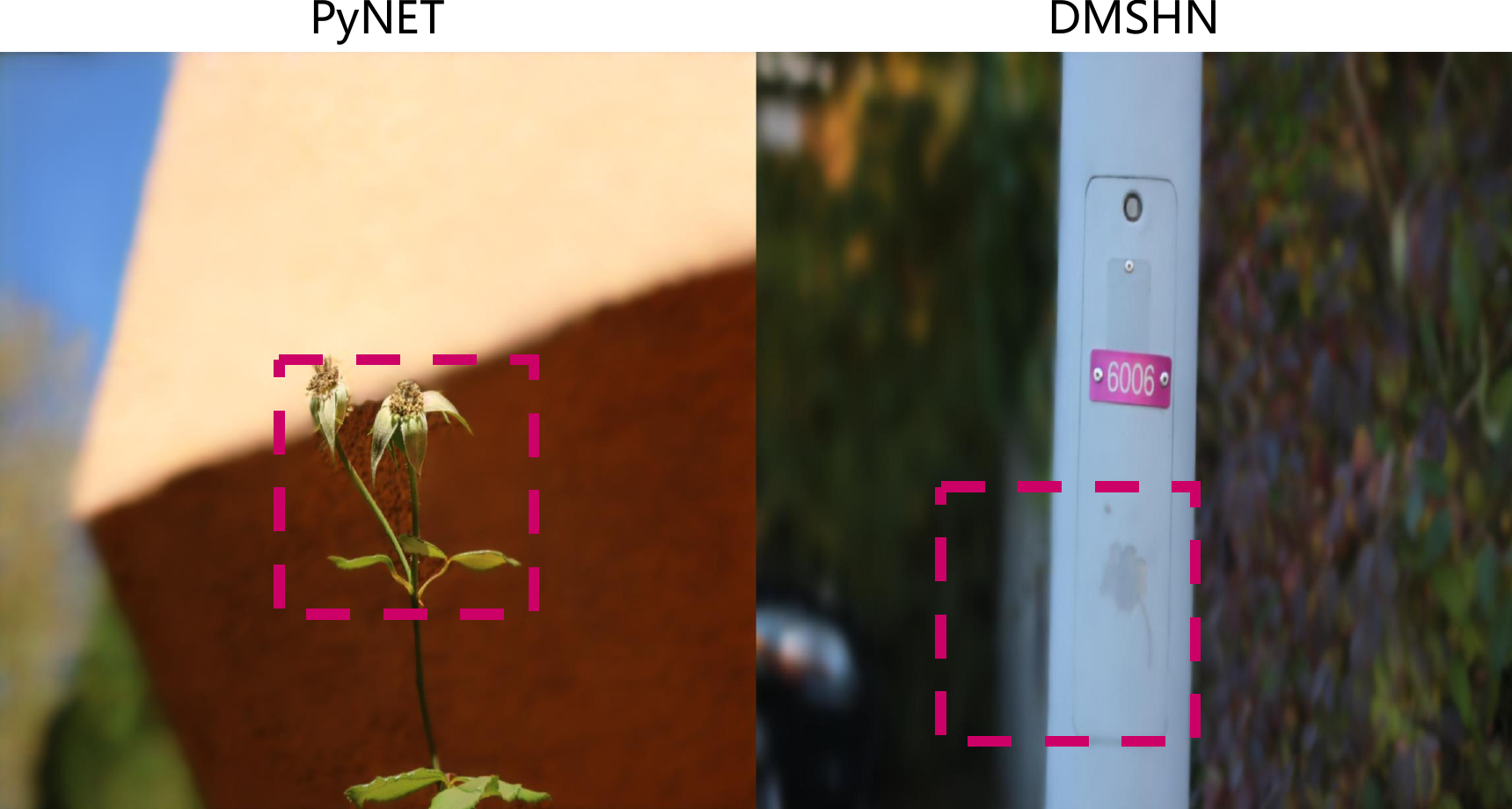}
\caption{There are noticeable artifacts from the bokeh rendering methods PyNET \cite{ignatov2020rendering} and DMSHN \cite{dutta2021stacked}.}
\label{fig:DMSHN_PyNET_flaw}
\end{figure}

\subsection{Motivation}\label{sec:mot}

Although existing bokeh effect rendering methods have achieved impressive photographic effects, the SOTA methods mainly adopt data-driven end-to-end networks with cumbersome architecture and complicated training strategies. For example, PyNET \cite{ignatov2020rendering} designs a seven-level UNet-like network and replaces the skip-connection with diversified convolution modules. They need to go through seven stages of progressive training, confirming model parameters layer by layer, to get the final rendering model. With respect to DMSHN \cite{dutta2021stacked}, the authors first design a three-level network with the encoder-decoder module on each level. Then they stack multiple three-level networks together to achieve the complete model. The training is a two-stage procedure with different losses (\ie, $L_1$ and SSIM loss in the first stage, MS-SSIM loss in the second stage). Even with such a complex design, the rendered bokeh images still exhibit errors somehow (see Fig.~\ref{fig:DMSHN_PyNET_flaw}).

Beyond the pure data-driven rendering methods, we think the hybrid schema which takes both the advantage of data-driven and physical-related priori is a more reasonable and promising approach. To this end, we refer to the physical process of bokeh rendering in normal sensors and try approximately simulating the process. We hope that the hybrid network design can help reduce the learning difficulty of the network, leading to a much lighter network and faster inference speed. With this hybrid solution, we are able to study the influence of bokeh-related physical parameters (\ie, depth-of-the-field) on the downstream visual intelligence tasks. In the following sections, we first introduce the physical model of bokeh rendering (Sec.~\ref{sec:camera_bokeh_modeling}) and then describe how we design the network to simulate it (Sec.~\ref{sec:Circle-of-confusion_Predictive_Network}).


\subsection{Bokeh Modeling with Sensors}\label{sec:camera_bokeh_modeling}


\subsubsection{Preliminary knowledge}
\begin{figure}
\centering
\includegraphics[width=0.8\linewidth]{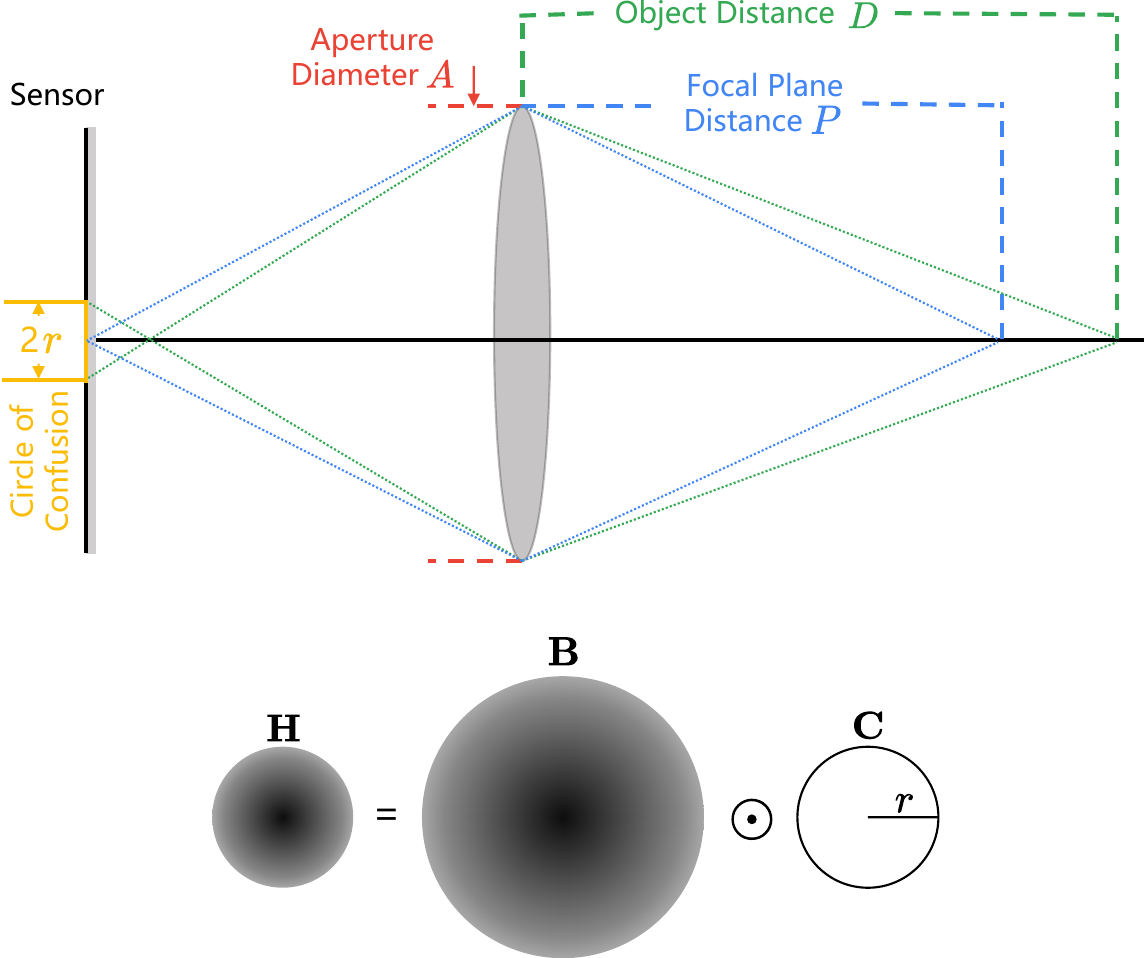}
\caption{Thin lens model and point spread function of Eq. \eqref{eq:radius_CoC} and \eqref{eq:PSF}.}
\label{fig:CoC_visual}
\end{figure}

Currently, the bokeh datasets are mainly captured by normal cameras. To simplify the optical calculation, we assume that the thickness of the lens is negligible. With the thin lens model, when a light source passes through the lens, it converges to a focal point on the image plane. If the light does not converge to a perfect focus, it will form a disk (\ie, circle of confusion (CoC)). The radius of CoC can be approximated by the following formula
\begin{align}\label{eq:radius_CoC}
r = \frac{A}{2} \times \frac{f}{D} \times \frac{\left| D-P \right|}{P-f},
\end{align}
which involves camera parameters focal length ($f$), distance of focal plane ($P$), aperture diameter ($A$), and object distance ($D$), \etc. For a given dataset captured by a certain type of camera, the camera parameters are fixed when taking the bokeh images, and only the object distance $D$ changes. Please see Fig.~\ref{fig:CoC_visual} for a pictorial illustration of Eq.~\eqref{eq:radius_CoC}

The point spread function (PSF) $\mathbf{H}$ of the view can be modeled as Hadamard product (\ie, element-wise multiplication) `$\odot$' of a 2D Butterworth filter $\mathbf{B}$ and a circular disk $\mathbf{C}$ \cite{abuolaim2021learning} with CoC's radius $r$,
\begin{align}\label{eq:PSF}
\mathbf{H} = \mathbf{B} \odot \mathbf{C}.
\end{align}
Here $\mathbf{H}$, $\mathbf{B}$, and $\mathbf{C}$ are all related to the object distance $D$. Please see Fig.~\ref{fig:CoC_visual} for a pictorial illustration of Eq.~\eqref{eq:PSF}. For each sublayer $\mathbf{I}_{m}$ that contains pixels from the same depth of the depth map \wrt the all-in-focus image $\mathbf{I}$, the blurred sublayer $\mathbf{I}'_{m}$ can be modeled as the convolution `$\ast$' of $\mathbf{I}_{m}$ and its PSF,
\begin{align}\label{eq:partial_blur}
\mathbf{I}'_{m} = \mathbf{I}_{m} \ast \mathbf{H}_{m}.
\end{align}
The bokeh image $\hat{\mathbf{I}}$ is the fusion of the blurred sublayers.

\subsubsection{Modeling}

\begin{figure}[]
\centering
\includegraphics[width=\linewidth]{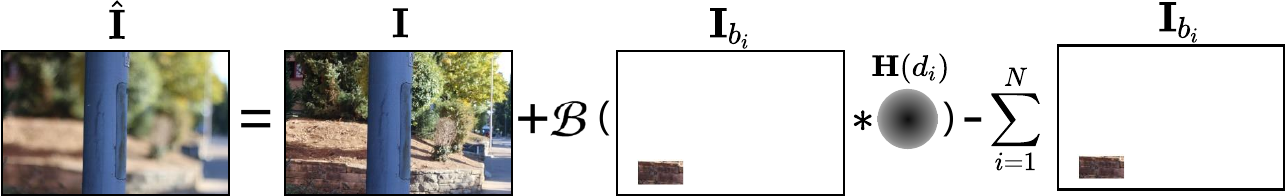}
\caption{The pictorial illustration of Eq.~\eqref{eq:blur_background_image}.}
\label{fig:aux_visual_eq7}
\end{figure}

Given an all-in-focus image $\mathbf{I}\in\mathbb{R}^{H\times W\times 3}$, the target is to obtain the bokeh image $\hat{\mathbf{I}}\in\mathbb{R}^{H\times W\times 3}$ corresponding to the all-in-focus image $\mathbf{I}$. Assume that the camera focuses on the foreground area and the all-in-focus image $\mathbf{I}$ consists of the foreground image $\mathbf{I}_f$ and background image $\mathbf{I}_b$ (\ie, $\mathbf{I} = \mathbf{I}_f+\mathbf{I}_b$). According to the property of bokeh (\ie, the foreground remains the same), the bokeh image $\hat{\mathbf{I}}$ consists of the foreground image $\mathbf{I}_f$ and the blurred background image $\mathbf{I}'_b$ (\ie, $\hat{\mathbf{I}} = \mathbf{I}_f + \mathbf{I}'_b$). Then we replace the $\mathbf{I}_f$ with $\mathbf{I}-\mathbf{I}_b$ and get
%
\begin{figure}[]
\centering
\includegraphics[width=\linewidth]{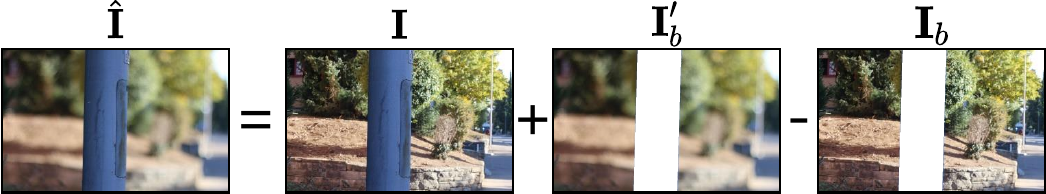}
\caption{The pictorial illustration of Eq.~\eqref{eq:bokeh_render}.}
\label{fig:aux_visual_eq4}
\end{figure}
\begin{align}\label{eq:bokeh_render}
\hat{\mathbf{I}} = \mathbf{I} +  \mathbf{I}'_b - \mathbf{I}_b,
\end{align}
impelling the model to learn the residual between $\mathbf{I}$ and $\hat{\mathbf{I}}$. Please see Fig.~\ref{fig:aux_visual_eq4} for a pictorial illustration of Eq.~\eqref{eq:bokeh_render}. For a sublayer $\mathbf{I}_{b_i}$ that contains pixels from the same depth (\ie, $d_i$ ($1 \leq i \leq N$, $N$ represents the number of the discrete depth layers in the depth map \wrt background image)) of the depth map \wrt background image $\mathbf{I}_b$. Set $\mathbf{M}(d_i)$ as a binary mask in which describes whether the pixels of depth $d_i$ are selected. $\mathbf{I}_{b_i}$ is the Hadamard product of $\mathbf{M}(d_i)\in\mathbb{R}^{H\times W}$ and $\mathbf{I}_b$, 
\begin{figure}[]
\centering
\includegraphics[width=0.7\linewidth]{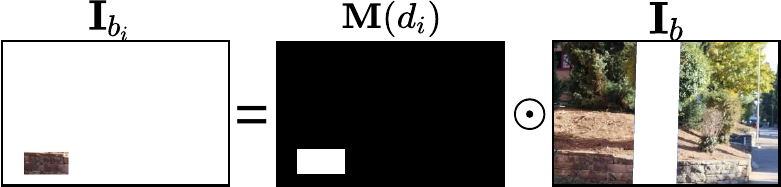}
\caption{The pictorial illustration of Eq.~\eqref{eq:background_image_sublayer}.}
\label{fig:aux_visual_eq5}
\end{figure}
\begin{align}\label{eq:background_image_sublayer}
\mathbf{I}_{b_i} = \mathbf{M}(d_i) \odot \mathbf{I}_b.
\end{align}
That is, $\mathbf{I}_b = \sum_{i=1}^N \mathbf{M}(d_i) \odot \mathbf{I}_b$. Please see Fig.~\ref{fig:aux_visual_eq5} for a pictorial illustration of Eq.~\eqref{eq:background_image_sublayer}. The blurred background image $\mathbf{I}'_b$ can be represented by blending (\ie, $\mathcal{B}(\cdot)$) the blurred sublayers $\mathbf{I}'_{b_i}$ ($\mathbf{I}'_{b_i}$ is the blurred version of $\mathbf{I}_{b_i}$) together. Then Eq. \eqref{eq:bokeh_render} can be rewritten as
\begin{align}\label{eq:decompose_background_image}
\hat{\mathbf{I}} = \mathbf{I} +  \mathcal{B}(\mathbf{I}'_{b_i}) - \sum_{i=1}^N \mathbf{I}_{b_i}.
\end{align}
According to Eq. \eqref{eq:PSF}, $\mathbf{I}'_{b_i}$ can be represented as the convolution of $\mathbf{I}_{b_i}$ and its corresponding PSF (\ie, $\mathbf{H}(d_i)$). Please see Fig.~\ref{fig:aux_visual_eq7} for a pictorial illustration of Eq.~\eqref{eq:blur_background_image}.
\begin{align}\label{eq:blur_background_image}
\hat{\mathbf{I}} &= \mathbf{I} + \mathcal{B}\big(\mathbf{I}_{b_i} \ast \mathbf{H}(d_i)\big) - \sum_{i=1}^N \mathbf{I}_{b_i}.
\end{align}
The function $\mathcal{B}(\cdot)$ depends on the processed masks $\mathbf{M}'(d_i)$ as the weight to fuse the blurred sublayers. The processed masks $\mathbf{M}'(d_i)$ are generated by the convolution of mask and its PSF (\ie, $\mathbf{M}(d_i) \ast \mathbf{H}(d_i)$) \cite{abuolaim2021learning}. Therefore,
\begin{figure*}
\centering
\includegraphics[width=\linewidth]{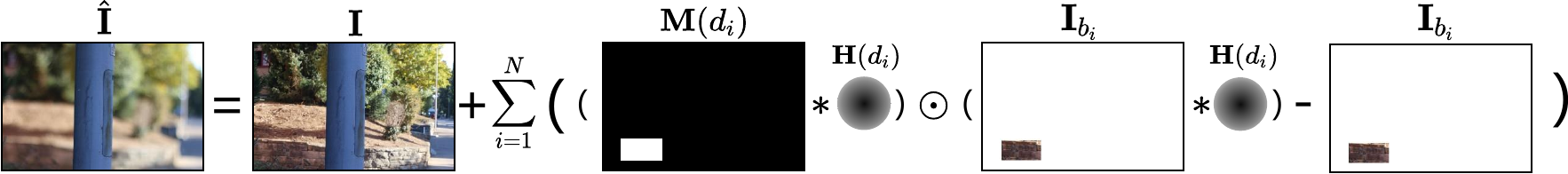}
\caption{The pictorial illustration of Eq.~\eqref{eq:unfold_blending}.}
\label{fig:Modeling_visual}
\end{figure*}
\begin{align}\label{eq:unfold_blending}
\hat{\mathbf{I}} = \mathbf{I} + \sum_{i=1}^N \big(\mathbf{M}(d_i) \ast \mathbf{H}(d_i)\big) \odot \big(\mathbf{I}_{b_i} \ast \mathbf{H}(d_i)\big) - \mathbf{I}_{b_i}.
\end{align}
In terms of a single pixel $\mathbf{I}_{x_0,y_0}$ (of which the depth is $d_0$) in the location [$x_0,y_0$] of $\mathbf{I}_b$, according to Eq. \eqref{eq:unfold_blending},
\begin{align}\label{eq:per_pixel_blur}
\hat{\mathbf{I}}_{x_0,y_0} &= \mathbf{I}_{x_0,y_0} + u(d_0) \times \mathbf{I}_{x_0,y_0} \ast \mathbf{H}(d_0) - \mathbf{I}_{x_0,y_0},
\end{align}
where $u(d_0)$ represents the value in the location [$x_0,y_0$] of the processed mask $\mathbf{M}'(d_0)$. $\mathbf{I}_{x_0,y_0}$ can be replaced by $u(d_0) \times \mathbf{I}_{x_0,y_0} \ast \frac{\mathbf{R}}{u(d_0)}$, where $\mathbf{R}$ is a kind of kernel which does not change the value of the pixel $\mathbf{I}_{x_0,y_0}$ and has the same size as $\mathbf{H}(d_0)$. Then we can rewrite the Eq. \eqref{eq:per_pixel_blur} into 
\begin{align}\label{eq:per_pixel_blur_simplify}
\hat{\mathbf{I}}_{x_0,y_0} = \mathbf{I}_{x_0,y_0} + u(d_0) \times \mathbf{I}_{x_0,y_0} \ast \mathbf{J}(d_0),
\end{align}
where $\mathbf{J}(d_0) = \mathbf{H}(d_0) - \frac{\mathbf{R}}{u(d_0)}$. Intuitively, we can find that for each pixel in the background image, its bokeh version is composed of its own value and a residual. The form of residual can be regarded as $ a \times b \ast c$. Obviously, the residual of pixels in the foreground image can also be written in this form (by simply setting $a$ to 0). Therefore, the bokeh rendering model can be summarized as the general format
\begin{align}\label{eq:whole_image_bokeh_simplify}
\hat{\mathbf{I}} = \mathbf{I}+ \sum_{i=1}^N \mathbf{v}(d_i) \odot \mathbf{I} \ast  \mathbf{L}(d_i),
\end{align}
where $\mathbf{v}(d_i)\in\mathbb{R}^{H\times W}$ is a weight map and $\mathbf{L}(d_i)$ means kernels (the number of $\mathbf{L}(d_i)$ is $H\times W$). The size of the kernels depends on the depth $d_i$.

If directly rendering the all-in-focus image with Eq. \eqref{eq:whole_image_bokeh_simplify}, there are two factors that lead to a huge computation problem, which is not convenient. \ding{182} The method needs to split the pixels into many sublayers (maybe up to several hundred) according to their depth. For each sublayer, a dedicated rendering operation is needed, which is time-consuming. \ding{183} For the pixels at deep depth, the size of the kernels is large, which makes convolution operation time-consuming.

To simplify the bokeh rendering, we solve the huge computation problem and improve the efficiency through approximation strategies with respect to the two factors. The details of the approximation and the architecture of our network are introduced in the next subsection.

\subsection{Circle-of-confusion Predictive Network}\label{sec:Circle-of-confusion_Predictive_Network}

\begin{figure*}
\centering
\includegraphics[width=0.75\linewidth]{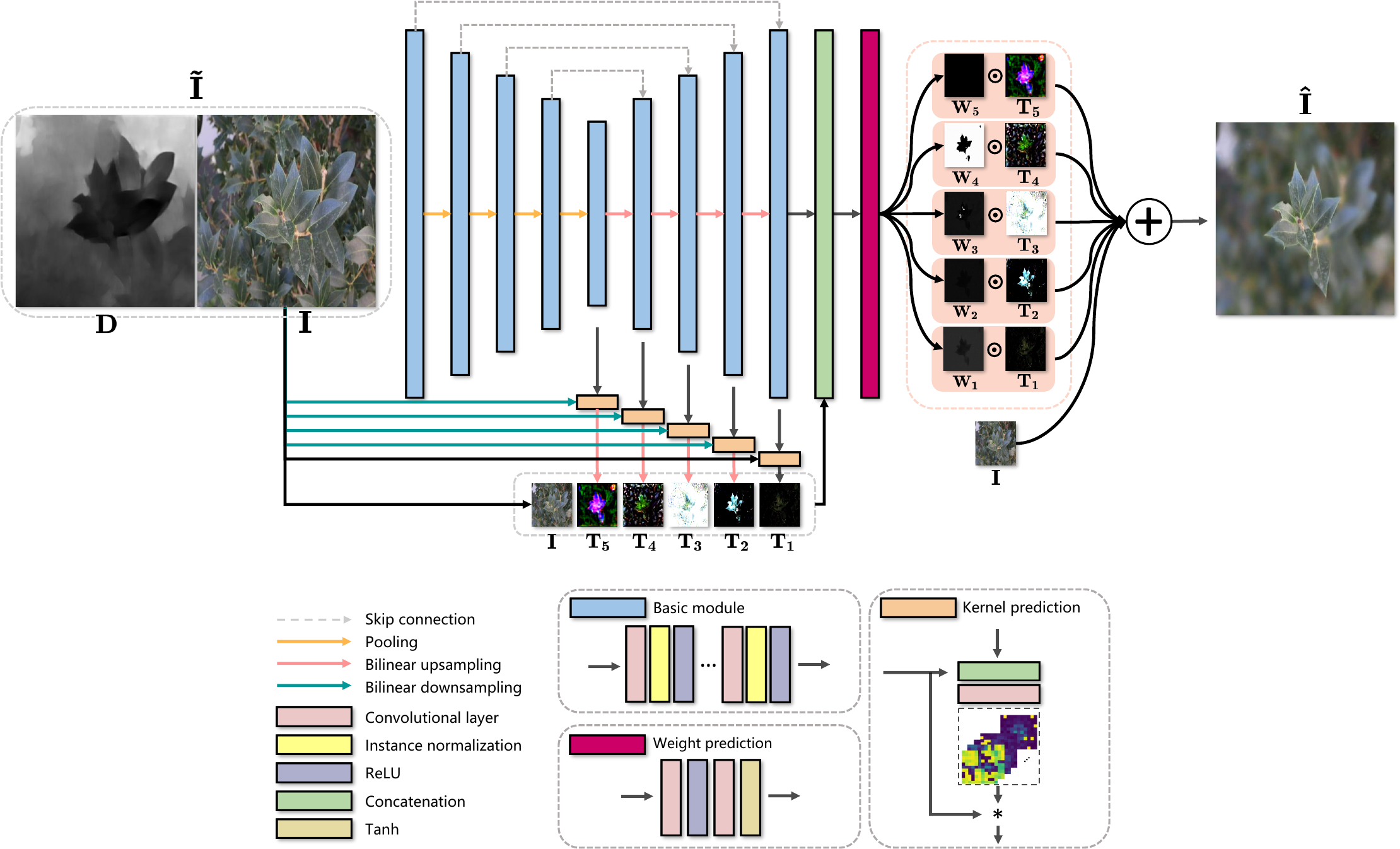}
\caption{Architectures and training pipeline of {\emph{circle-of-confusion predictive network (CoCNet)}}. Please note the depth map $\mathbf{D}$ is optional.}
\label{fig:archs}
\end{figure*}
\subsubsection{Approximation strategy.} In pursuit of operating efficiency, we approximate the physical principle of bokeh rendering by simplifications.

For the problem of having too many sub-layers and large convolution kernel size, we propose the following solutions respectively. To reduce the number of sublayers in the calculation, we propose to use finite templates to approximate the effect, as shown in Eq. \eqref{eq:whole_image_bokeh_simplify_approximate}. Here we utilize the idea of sparse coding to simulate a complex image by weighted fusion of a few templates. Please note that the formula form is the same as the Eq. \eqref{eq:whole_image_bokeh_simplify}, which guarantees the physical correlation of the approximation formula.
\begin{align}\label{eq:whole_image_bokeh_simplify_approximate}
\hat{\mathbf{I}} = \mathbf{I}+ \sum_{j=1}^Q \mathbf{w}'_j \odot \mathbf{I} \ast  \mathbf{K}'_j.
\end{align}
The $\mathbf{w}'_j\in\mathbb{R}^{H\times W}$ is a weight map and $\mathbf{K}'_j$ is a group of kernels. $Q$ is the number of templates. In this paper, it is 5.

With regard to large kernel size, it is apparent that we inevitably need large convolution kernels to simulate the blur effect of distant objects. Thus we use the following formula to approximate Eq. \eqref{eq:whole_image_bokeh_simplify_approximate}. The $\mathbf{w}_j\in\mathbb{R}^{H\times W}$ is a weight map and $\mathbf{K}_j$ is a group of kernels.
\begin{align}\label{eq:whole_image_bokeh_simplify_approximate_twice}
\hat{\mathbf{I}} = \mathbf{I}+ \sum_{j=1}^Q \mathbf{w}_j \odot \mathcal{U}_j\big(\mathcal{D}_j(I) \ast \mathbf{K}_j\big),
\end{align}
where $\mathcal{U}_j(\cdot)$ and $\mathcal{D}_j(\cdot)$ are upsampling/downsampling method. They will upsample/downsample the width/height of an image by $2^j$. The main idea is to apply convolution kernels of the same size on images of different scales. Then, by scaling the images back to a uniform size with different ratios, the blur effect of different convolution kernel sizes can be simulated.

\subsubsection{Architecture of CoCNet}

To implement Eq. \eqref{eq:whole_image_bokeh_simplify_approximate_twice}, we design a network (\ie, $\mathcal{T}(\cdot)$) which is based on an encoder-decoder architecture, as shown in Fig. \ref{fig:archs}. Since the Eq. \eqref{eq:whole_image_bokeh_simplify_approximate_twice} implicitly predicts circle-of-fusion for each pixel, thus we call the network circle-of-confusion predictive network (CoCNet).
\begin{align}\label{eq:CoCNet_function}
\hat{\mathbf{I}} = \mathcal{T}(\tilde{\mathbf{I}}),
\end{align}
where $\tilde{\mathbf{I}}$ is the input of the network. It includes the all-in-focus image $\mathbf{I}\in\mathbb{R}^{H\times W\times 3}$ and the depth map $\mathbf{D}_{|[0,1]}\in\mathbb{R}^{H\times W}$ of $\mathbf{I}$. Please note that the depth map is optional. As shown in Table~\ref{Table:bokeh_quantitative}, the network without the depth map can also achieve state-of-the-art performance. The reason why we take it as the input is to maintain the formula consistency of natural/adversarial bokeh rendering methods. The encoder-decoder network consists of $2Q-1$ basic module layers, in which the encoder has $Q-1$ layers (from $\mathbf{Enc}_1$ to $\mathbf{Enc}_{Q-1}$) and the decoder has $Q$ layers (from $\mathbf{Dec}_Q$ to $\mathbf{Dec}_{1}$). The encoder layers are connected by downsampling and the decoder layers are connected by upsampling. There are skip connections between the encoder layer $j$ and decoder layer $j$ ($1 \leq j \leq Q-1$). We assume that the feature map outputs of decoder layers are $\mathbf{Dec}_j(\mathbf{\tilde{\mathbf{I}}})$. The template $\mathbf{T}_j\in\mathbb{R}^{H\times W\times 3}$ is
\begin{align}\label{eq:template_generation}
\mathbf{T}_j = \mathcal{U}_j\Big(\mathcal{D}_j(\mathbf{I}) \ast \varphi\big(\mathbf{De}_{Q-j+1}(\tilde{\mathbf{I}})\big)\Big),
\end{align}
where $\varphi(\cdot)$ represents the kernel prediction module. It will output $\frac{H}{2^j} \times \frac{W}{2^j}$ kernels, the width/height of each kernel is $k$. Please note that here we assume the number $\frac{H}{2^j}$ and $\frac{W}{2^j}$ are integers. For decimals, we can adjust the scaling flexibly.

We can achieve the weight map of each template by 
\begin{align}\label{eq:weight_generation}
\mathbf{w}_j =  \mathcal{W}\big([\mathbf{I}, T_1, T_2, \cdots, T_Q, \mathbf{Dec}_1(\tilde{\mathbf{I}})]\big),
\end{align}
where $\mathcal{W}(\cdot)$ is the weight prediction module. Finally, the bokeh image $\hat{\mathbf{I}}$ can be achieved by
\begin{align}\label{eq:CoCNet_function_final}
\hat{\mathbf{I}} = \mathbf{I} + \sum_{j=1}^Q \mathbf{w}_j \odot \mathbf{T}_j.
\end{align}
DBSI \cite{dutta2021depth} may look similar to our method since it also fuses the all-in-focus image with templates (\ie, Gaussian-blurred versions of the all-in-focus image). However, our method is fundamentally different from their approach since they do not follow the physical model. The Gaussian blur is physically different from the bokeh effect and they prepare the Gaussian-blurred version of the all-in-focus image manually. To be specific, they generate several Gaussian-blurred images of an all-in-focus image as templates. The kernel sizes of the Gaussian blur are manually set, which is cumbersome and lacks generality. Furthermore, the preparation procedure for the templates is time-consuming since they have to generate these Gaussian-blurred templates for each all-in-focus image. In contrast, the templates in our method are physical-aware and generated by the network automatically.\\
\noindent\textbf{Limitation.}
Since our method generates templates to simulate the bokeh effect, the network may inevitably generate the bokeh effect, even for objects that are located close to the all-in-focus physical range. This is a potential limitation of our method.

\section{Natural \& Adversarial Bokeh on Visual Task}\label{sec:influence_on_visual_task}
Since the CoCNet follows the physical priori, it gives us the ability to study the influence of the natural bokeh effect on visual tasks. In Sec.~\ref{sec:Influence_of_Natural_Bokeh}, we study the influence of bokeh on image classification as an example. Furthermore, the bokeh effect may be maliciously exploited to attack visual tasks and cause harm to society. Thus in Sec.~\ref{sec:Unrestricted_Adversarial_Bokeh_Attack}, we further propose three kinds of depth-guided adversarial bokeh attack methods as unrestricted attacks to reveal the extent of the harm.

\subsection{Influence of Natural Bokeh}\label{sec:Influence_of_Natural_Bokeh}
Image degradations may drop the performance of neural networks on the visual tasks \cite{endo2020classifying,thao2021nonuniform,pei2019effects}. Although the bokeh effect aims to produce aesthetically pleasing photos, it may come with some unexpected bad influences on visual tasks. Whether and how much it drops the performance of neural networks is an interesting and meaningful problem. Here we take the classical visual task (\ie, image classification) as the target to show the influence of the natural bokeh effect.

To study the influence of the natural bokeh effect, there are two challenges. First, there are not enough all-in-focus and bokeh image pairs (\ie, enough data) to support the study. Second, the existing bokeh datasets are not designed for classification tasks, thus leading to the lack of classification labels. To solve these two problems, we need a bokeh rendering method that can generate an unlimited number of natural bokeh images from the labeled all-in-focus images. Since the CoCNet follows the physical priori and the generated bokeh images reflect the characteristics of natural bokeh, it is obvious that the proposed CoCNet is the rescue. 

We conduct the experiments on the NeurIPS-2017 DEV dataset (\ie, an ImageNet-like dataset) and on four classical CNNs (\ie, ResNet50, VGG, DenseNet, and MobileNetV2). As shown in Table~\ref{Table:bokeh_attack}, ``aif'' and ``bokeh'' in each group mean the all-in-focus images and corresponding bokeh images. We can find that the bokeh effect drops the classification performance of all four CNNs by an average of 7\%. This demonstrates the minor aggressiveness of the bokeh effect.

\subsection{Depth-guided Adversarial Bokeh Attack}\label{sec:Unrestricted_Adversarial_Bokeh_Attack}

Although natural bokeh has minor adverse effects on neural networks, some people may maliciously generate natural adversarial examples without any human-noticeable noise artifacts to attack the perception model. To fully study the influence of adversarial bokeh rendering, We further propose three kinds of depth-guided adversarial bokeh attacks with CoCNet and reveal the vulnerability of neural networks.

\ding{182} In traditional restricted adversarial attacks, the methods usually add attack noise to the image. However, directly adding noise to the bokeh image violates the smoothness feature of the bokeh, which may arouse suspicion and raise the defense of the detection mechanism. It is better to apply unrestricted and non-suspicious attacks on the bokeh image by modifying the input of CoCNet (\ie, $\tilde{\mathbf{I}}$). Then the calculated deceptive information is embedded into the aesthetical bokeh effect generation procedure, which is more dangerous to neural networks and more worthy of attention. In our attack method, we choose to modify the physical factors (\ie, depth map) rather than the all-in-focus image to simulate the adversarial bokeh attack, which can better construct the relationship between the real-world physical factors and bokeh aggression. 

Given a pretrained CNN (\ie, $\phi(\cdot)$) for image classification task, an all-in-focus image $\mathbf{I}$ and its depth map $\mathbf{D}$. We first propose to simply apply gradient-based attack (\ie, ``gda'') to achieve attacked depth map $\mathbf{D}^{*} = \mathbf{D} + \delta$ by optimizing the objective function 
\begin{align}\label{eq:pgd}
\delta = \argmax_{\delta^{*}} \mathcal{J} \Big(\phi\big(\mathcal{T}([\mathbf{I},\mathbf{D} + \delta^{*}])\big),y \Big), \nonumber \\
~\text{subject to}~\|\delta^{*}\|_p \leq \epsilon,
\end{align}
\noindent
where $\mathcal{J}(\cdot)$ denotes the cross-entropy loss function with $y$ being the annotation of the all-in-focus image $\mathbf{I}$. By directly applying a gradient-based attack on the depth map, although the depth map is with noise, the rendered bokeh image looks natural. The method avoids adding suspicious noise to the bokeh image by transforming the noise in the depth map into the inconspicuous bokeh effect.

\ding{183} Although the noise in the depth map is hidden and imperceptible during the bokeh generation process, we propose a smooth gradient-based attack method (\ie, ``sm-gda'') to further maintain the smoothness property of the depth map, which will lead to a more natural bokeh effect, as shown in Table~\ref{Table:bokeh_attack}. In specific, we change the value of gradient $\nabla_{\mathbf{D}_{i,j}}$ ($1\leq i \leq H, 1\leq j \leq W$) of depth map $\mathbf{D}\in\mathbb{R}^{H\times W}$ according to its neighbors. The number of neighbors used for reference is dependent on $l$, where $l$ is the kernel size of the smooth function $\mathcal{S}(\cdot)$. In detail, the value $\nabla_{\mathbf{D}_{i,j}}$ is changed according to $\nabla_{\mathbf{D}_{i\pm(l/2),j\pm(l/2)}}$. We obtain the attacked depth map $\mathbf{D}^{*} = \mathbf{D} + \mathcal{S}(\delta,l)$ with the Eq.~\eqref{eq:pgd}. 

\ding{184} Furthermore, applying a gradient-based attack on the depth map ignores an important characteristic of the bokeh image (\ie, the object in the focus region is clear). With an intuitive idea that the variation in the focus region attracts more attention than variation in a blurred region, we aim to remove the variation in the focus region and only apply a gradient-based attack on the background. We call this background-guided gradient-based attack (\ie, ``bg-gda''). 
We obtain the attacked depth map $\mathbf{D}^{*} = \mathbf{D} + \delta^{*} \odot \mathbf{BR}$ with the same objective as in Eq. \eqref{eq:pgd}. $\mathbf{BR}$ is the background region (represented by a binary map) obtained by applying threshold on the depth map.


\begin{table*}
\centering
\caption{Comparison between \textbf{CoCNet} and other SOTA bokeh effect rendering methods. The ablation study results are on the right. 
}
\setlength{\tabcolsep}{3.5pt}
\resizebox{\linewidth}{!}{
\begin{tabular}{l|cccccc||cccccc}
\hline 
 & MPFNet & BRViT & PyNET & DMSHN & DMSHN-os & CoCNet-b3 & CoCNet-b2 & CoCNet-b3-less & CoCNet-b2-less & CoCNet-b3-$L_2$ & CoCNet-b3-dir & CoCNet-b3-depth \tabularnewline
\hline 
PSNR $\uparrow$& 24.74 & 24.76 & 24.21 & 24.73 & 24.57 & \textbf{24.78} & 24.70 & 24.72 & 24.62 & 24.63 & 24.24 & 24.81\tabularnewline
SSIM $\uparrow$& 0.8806 & \textbf{0.8904} & 0.8593 & 0.8619 & 0.8558 & 0.8604 & 0.8603 & 0.8597 & 0.8579 & 0.8611 & 0.8507 & 0.8607\tabularnewline
LPIPS $\downarrow$& 0.0596 & \textbf{0.0509} & 0.0693 & 0.0601 & 0.0631 & 0.0625 & 0.0632 & 0.0635 & 0.0630 & 0.0639 & 0.0686 & 0.0624\tabularnewline
Para.(M) $\downarrow$& 6.12 & 123.14 & 47.5 & 10.84 & 10.84 & 7.34 & 5.18 & 1.91 & 1.37 & 7.34 & 7.07 & 7.61\tabularnewline
\hline 
\end{tabular}}
\label{Table:bokeh_quantitative}
\end{table*}

\begin{figure*}[tbp]
\centering
\includegraphics[width=0.9\linewidth]{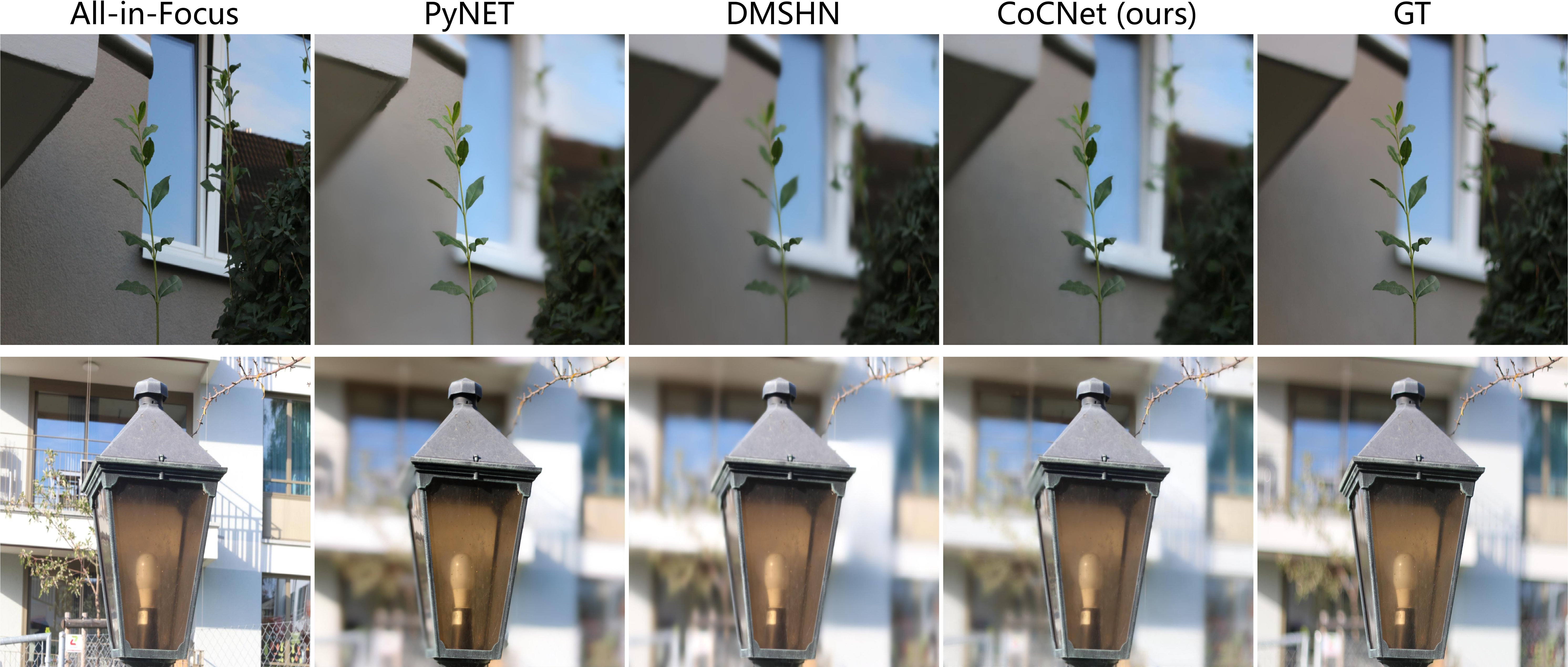}
\caption{Bokeh effect comparison. Our method has comparable results with the state-of-the-art bokeh rendering methods PyNET \cite{ignatov2020rendering} and DMSHN \cite{dutta2021stacked}.}
\label{fig:bokeh_comparison}
\end{figure*}

\section{Experiments}\label{sec:exp}
We verify the effectiveness of the method from three aspects: the bokeh effect and efficiency (Sec.~\ref{exp:bokeh_effect}), the attack success rate (Sec.~\ref{exp:adversarial_attack}), and the improvement to downstream tasks (Sec.~\ref{exp:down_stream_task}).

\subsection{Target Model}
We take four SOTA bokeh effect rendering methods (\ie, DMSHN \cite{dutta2021stacked}, PyNET \cite{ignatov2020rendering}, MPFNet \cite{wang2022self}, and BRViT \cite{nagasubramaniam2022bokeh}) as the baseline for bokeh effect comparison. The advantage of DMSHN is that the model is small enough to run on consumer-level phones (\eg, Qualcomm Snapdragon 855+ processor, Adreno 640 GPU and 8GB RAM) with several seconds for an image. Though the DMSHN model is small, our model is smaller, which is introduced in Sec.~\ref{exp:bokeh_effect}. To evaluate the influence of adversarial bokeh examples on CNNs, we take ResNet50, VGG, MobileNetV2, and DenseNet as the target model. For downstream tasks, we aim to improve the performance of the SOTA defocus deblurring method \cite{Lee_2021_CVPR} with the adversarial bokeh examples. 

\subsection{Experimental Setup}\label{sec:setting}
\paragraph{Datasets.}
There are several datasets involved in our experiment. Ignatov \etal \cite{ignatov2020rendering} open-source the ``Everything is Better with Bokeh!'' (EBB!) dataset which was used in AIM 2020 Bokeh Effect Synthesis Challenge. The dataset contains 4,694 pairs of bokeh and bokeh-free images captured using a narrow aperture (f/16) and a high aperture (f/1.8). The resolution of images is around 1024 $\times$ 1536 pixels. We resized the images to 1024 $\times$ 1024 in our experiment. This dataset is used in the training (4,400) and testing (294) of the bokeh effect rendering network. We use the NeurIPS-2017 DEV dataset \cite{kurakin2018adversarial} as the testing dataset for the classification task, which contains 1,000 ImageNet-like images and is used by NIPS 2017 adversarial attacks and defenses competition. For downstream tasks, we use the dual-pixel defocus deblurring (DPDD) \cite{abuolaim2020defocus} dataset. The DPDD dataset provides 500 dual-pixel images captured by a Canon EOS 5D Mark IV. 

\paragraph{Metrics.}
To measure the similarity between prediction and ground truth bokeh images, we use peak signal-to-noise ratio (PSNR) \cite{wang2004image}, structural similarity (SSIM) \cite{hore2010image} and learned perceptual image patch similarity (LPIPS) \cite{zhang2018unreasonable} as the metric. PSNR is the most commonly used measurement for the reconstruction quality of lossy compression. SSIM is used for measuring the similarity between two images. LPIPS is a metric that uses the features of neural networks to judge the similarity of images. Higher means more different and lower means more similar. We also use model parameters (Para.(M), M means million) as the metric to evaluate the model size.

\paragraph{Implementation details.}
In CoCNet, as shown in Fig. \ref{fig:archs}, we set the layer numbers of the encoder-decoder to be 9 and the conv kernels in the encoder-decoder of CoCNet are of size 3 with striding 1 and padding 1. The group (\ie, convolutional layer, instance normalization, ReLU) numbers in the basic module are 3 and the kernel size of the kernels in the kernel prediction module is (11,11). We use the Adam optimizer with $2e^{-4}$ learning rate and the batch size is 2. We also use Xavier initialization \cite{glorot2010understanding} on the weight of the network. The loss function used by us is $L_1$ and $SSIM$ loss, $L = L_1 + (1- L_{SSIM})$. 
All the experiments were run on an Ubuntu 16.04 system with an Intel(R) Xeon(R) CPU E5-2699 with 196 GB of RAM, with an NVIDIA Tesla V100 GPU of 32G RAM.  

\subsection{Bokeh Effect}\label{exp:bokeh_effect}
\paragraph{Quantitative and qualitative evaluation.}
In this section, we compare our method to the current four
state-of-the-art bokeh rendering methods (\ie, PyNET \cite{ignatov2020rendering}, DMSHN \cite{dutta2021stacked}, MPFNet \cite{wang2022self} and BRViT \cite{nagasubramaniam2022bokeh}) that were tuned specifically for the bokeh rendering. 
The quantitative results on comparing with state-of-the-art results and the ablation study are shown in Table~\ref{Table:bokeh_quantitative}.
In the first row, the ``PyNET'', ``DMSHN'', ``MPFNet'' and ``BRViT'' represent the official best-pretrained models respectively. ``DMSHN-os'' means using our training strategy, that is, training the DMSHN model with $L_1$ and $L_{SSIM}$ loss in one stage. Because the models are trained without a depth map as input, we train ``CoCNet-b3'', a model only using the all-in-focus image as input to keep the input information consistent with them. We find that ``CoCNet-b3'' has already achieved comparable performance with the state-of-the-art methods. From the table, we can find that the performance of CoCNet and DMSHN are significantly better than PyNET (higher similarity and fewer parameters). Compared with ``DMSHN-os'', our model is better on all four metrics. Compared with the official model ``DMSHN'', our model has comparable performance with fewer parameters. From the paper of DMSHN \cite{dutta2021stacked}, we can find that the PSNR result of PyNET evaluated by them is 24.93. However, we have to claim that, there is a nonrigorous setting in the DMSHN paper that makes the PSNR result of PyNET in the DMSHN paper to be higher than it actually should be. The problem is from the dataset setting. The EBB! dataset has 4694, 200, and 200 pairs of images in its training, validation and test set. Since the ground truth images of validation and test set are not available yet, DMSHN uses part of the image pairs (294 pairs) of the training set as the test set. This means, in the setting of DMSHN, the training set has 4400 (4694-294=4400) pairs of images and the test set has 294 pairs of images. However, in the setting of PyNET paper, they use all the 4694 pairs of images as the training set. Thus testing the PyNET on the test set of DMSHN is not fair since the PyNET has seen the test set of DMSHN in its training set. This is the reason why the PSNR result claimed in the DMSHN paper (they use the model pretrained by PyNET) is higher than the official value claimed by PyNET itself. In contrast, we additionally train the PyNET by ourselves and avoid the above unfair setting. For MPFNet, our CoCNet has similar metric values to it. For BRViT, our method is a bit worse in image quality but much smaller in model size. Please note that since MPFNet and BRViT do not provide pretrained model and codes respectively, the results are referred to their paper.

\begin{table*}
\centering
\setlength{\tabcolsep}{2pt}
\caption{Performance of the attack method on four models and the NeurIPS-2017 DEV dataset. The similarity between the attacked bokeh images and the all-in-focus images. ``aif'' means all-in-focus image. ``bokeh'' means bokeh image generated by CoCNet. ``sm-gda'', ``gda'' and ``bg-gda'' mean smooth gradient-based attacked bokeh image, gradient-based attacked bokeh image and background-guided gradient-based attacked bokeh image, respectively. The \textbf{\color{black}top-1} value of each metric in each model is bolded.}
\resizebox{0.9\linewidth}{!}{
\begin{tabular}{l|ccccc|ccccc|ccccc|ccccc}
\toprule 
 & \multicolumn{5}{c|}{\textbf{\emph{ResNet50}}} & \multicolumn{5}{c|}{\textbf{\emph{VGG}}} & \multicolumn{5}{c|}{\textbf{\emph{DenseNet}}} & \multicolumn{5}{c}{\textbf{\emph{MobileNetV2}}}\tabularnewline
 & aif & bokeh & gda & sm-gda & bg-gda & aif & bokeh & gda & sm-gda & bg-gda & aif & bokeh & gda & sm-gda & bg-gda & aif & bokeh & gda & sm-gda & bg-gda \tabularnewline
\midrule
Acc. $\downarrow$& 0.923 & 0.864 & 0.046 & 0.054 & 0.038 & 0.890 & 0.807 & 0.045 & 0.048 & 0.041 & 0.946 & 0.885 & 0.048 & 0.060 & 0.038 & 0.885 & 0.797 & 0.024 & 0.038 & 0.020\tabularnewline
PSNR $\uparrow$& \textbackslash{} & \textbackslash{} & 26.560 & \textbf{26.597} & 26.505 & \textbackslash{} & \textbackslash{} & 26.482 & \textbf{26.556} & 26.378 & \textbackslash{} & \textbackslash{} & 26.547 & \textbf{26.582} & 26.472 & \textbackslash{} & \textbackslash{} & 26.550 & \textbf{26.573} & 26.484 \tabularnewline
SSIM $\uparrow$& \textbackslash{} & \textbackslash{} & \textbf{0.8846} & 0.8842 & 0.8832 & \textbackslash{} & \textbackslash{} & 0.8830 & \textbf{0.8850} & 0.8813 & \textbackslash{} & \textbackslash{} & 0.8842 & \textbf{0.8848} & 0.8828 & \textbackslash{} & \textbackslash{} & 0.8847 & \textbf{0.8846} & 0.8832 \tabularnewline
LPIPS $\downarrow$& \textbackslash{} & \textbackslash{} & 0.0307 & \textbf{0.0302} & 0.0313 & \textbackslash{} & \textbackslash{} & 0.0318 & \textbf{0.0302} & 0.0327 & \textbackslash{} & \textbackslash{} & 0.0314 & \textbf{0.0303} & 0.0322 & \textbackslash{} & \textbackslash{} & 0.0305 & \textbf{0.0300} & 0.0311 \tabularnewline
\bottomrule 
\end{tabular}}
\label{Table:bokeh_attack}
\end{table*}
\begin{table*}
\centering
\caption{Attack transferability evaluation. The images are obtained by attacking one of the four models (\ie,ResNet50, VGG, DenseNet and MobileNetV2) and testing on the other three. A lower value means better transferability.}
\setlength{\tabcolsep}{2pt}
\resizebox{0.8\linewidth}{!}{
\begin{tabular}{l|ccc|ccc|ccc|ccc}
\toprule 
 & \multicolumn{3}{c|}{\textbf{\emph{ResNet50}}} & \multicolumn{3}{c|}{\textbf{\emph{VGG}}} & \multicolumn{3}{c|}{\textbf{\emph{DenseNet}}} & \multicolumn{3}{c}{\textbf{\emph{MobileNetV2}}}\tabularnewline
\cline{2-13} \cline{3-13} \cline{4-13} \cline{5-13} \cline{6-13} \cline{7-13} \cline{8-13} \cline{9-13} \cline{10-13} \cline{11-13} \cline{12-13} \cline{13-13} 
 Acc. $\downarrow$ & VGG & DenseNet & MobileNetV2 & ResNet50 & DenseNet & MobileNetV2 & ResNet50 & VGG & MobileNetV2 & ResNet50 & VGG & DenseNet\tabularnewline
\hline 
aif & 0.890 & 0.946 & 0.885 & 0.923 & 0.946 & 0.885 & 0.923 & 0.890 & 0.885 & 0.923 & 0.890 & 0.946\tabularnewline
bokeh & 0.807 & 0.885 & 0.797 & 0.864 & 0.885 & 0.797 & 0.864 & 0.807 & 0.797 & 0.864 & 0.807 & 0.885\tabularnewline
gda & 0.695 & 0.771 & 0.675 & 0.754 & 0.796 & 0.664 & 0.712 & 0.691 & 0.675 & 0.766 & 0.676 & 0.803\tabularnewline
sm-gda & 0.714 & 0.798 & 0.717 & 0.789 & 0.832 & 0.711 & 0.759 & 0.714 & 0.720 & 0.786 & 0.704 & 0.831\tabularnewline
bg-gda & \textbf{0.684} & \textbf{0.744} & \textbf{0.669} & \textbf{0.731} & \textbf{0.789} & \textbf{0.648} & \textbf{0.684} & \textbf{0.684} & \textbf{0.668} & \textbf{0.758} & \textbf{0.665} & \textbf{0.800}\tabularnewline
\bottomrule 
\end{tabular}
}
\label{Table:attack_transferability}
\end{table*}

\textit{Ablation study.}
To study whether the model can be more lightweight, we adjust the optional variables of CoCNet to comprehensively demonstrate the model. \ding{182} The ``CoCNet-b3'' uses 3 groups in the basic module, thus we reduce it to 2 to see the influence on performance (\ie, ``CoCNet-b2'').  We can find that it has a close performance to ``DMSHN'' with half of the model size. \ding{183} We also try to reduce the kernel number in the convolutional layers of the basic module to half, obtaining ``CoCNet-b3-less'' and ``CoCNet-b2-less''. We can find that the ``CoCNet-b3-less'' model has a close performance to ``DMSHN'' with only $\frac{1}{5}$ model size, which is extremely small. With respect to ``CoCNet-b2-less'', though its model size is the smallest in the table, its performance has an obvious gap to ``DMSHN''. According to this discovery, we plan to improve the model which has less than 2M parameters in future work.

We also try different loss functions, network architecture and input. \ding{184} For loss function, we try $L_2+(1-L_{SSIM})$ loss, named ``CoCNet-b3-$L_2$''. It is not as good as using $L_1+(1-L_{SSIM})$ loss. \ding{185} Furthermore, to verify the function of the kernel prediction module (used to simulate the priori of convolution operation in Eq. \eqref{eq:whole_image_bokeh_simplify_approximate_twice}), we directly concatenate the feature outputs of each decoder layer and the corresponding downsampled all-in-focus image, adding convolution layer to process them and obtain templates. This operation takes the same input as the kernel prediction module and also output templates. The main difference is the way processing input. That is, the kernel prediction module applies convolution on the downsampled all-in-focus image with the predicted kernel while this operation (similar to the operation in ``PyNET'' and ``DMSHN'') only takes the downsampled all-in-focus image as information and hopes the network can learn the bokeh effect by itself. The model is named ``CoCNet-b3-dir''. We can find that the performance is far away from using the priori-based architecture, which points out the effectiveness of priori. \ding{186} Although our network can learn the depth-related weight map and templates by itself, adding depth maps as input provides information more directly. This is why we add the depth map as input when introducing the network architecture in Sec.~\ref{sec:Circle-of-confusion_Predictive_Network}. Furthermore, the adversarial bokeh attack method also needs to change the depth map. Thus we train the model with an estimated depth map as the complete model. The depth map is generated by Deeplens \cite{wang2018deeplens} and the corresponding model is ``CoCNet-b3-depth''. We achieve the same conclusion as \cite{dutta2021stacked}, that is, the depth map boosts minimal to the performance. Considering the time decay in generating depth maps, it is not suitable for consumer-level phones.

\subsection{Failure Samples}\label{exp:failure_samples}
\begin{figure}
\centering
\includegraphics[width=0.9\linewidth]{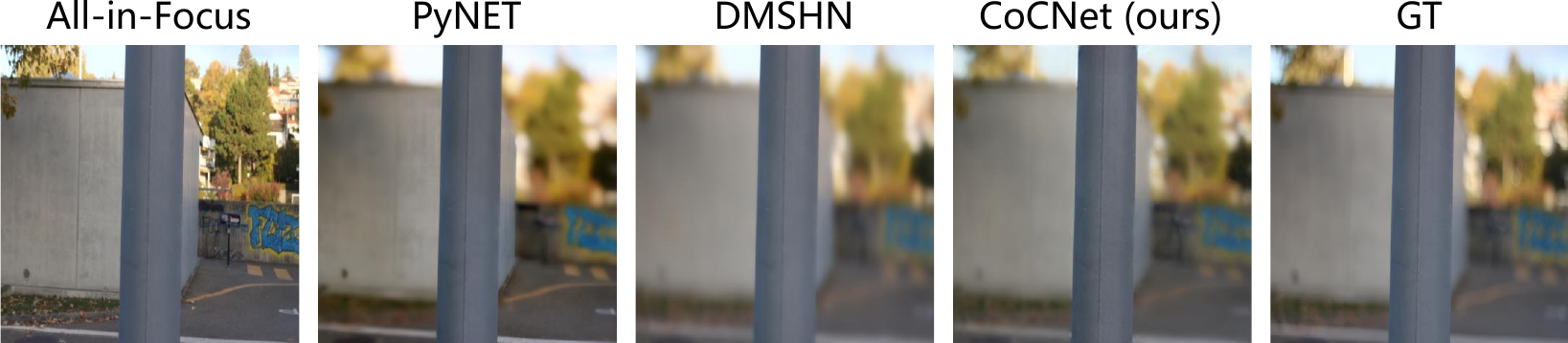}
\caption{Failure samples of CoCNet, DMSHN and PyNET.}
\label{fig:failure_samples}
\end{figure}

We carefully observe the bokeh image generated by CoCNet and find it is poor at generating the bokeh effect of spots. Please see Fig.~\ref{fig:failure_samples}, we can find that the CoCNet, DMSHN and PyNET are all poor at generating the bokeh effect of small spots, which needs further research.

\subsection{Adversarial Attack Accuracy}\label{exp:adversarial_attack}
We apply gradient-based attack, smooth gradient-based attack and background-guided gradient-based attack on four pretrained models (ResNet50, VGG, MobileNetV2 and DenseNet). As shown in Table~\ref{Table:bokeh_attack}, we demonstrate the accuracy of each model to the all-in-focus images (\ie, ``aif'' for short), bokeh image generated by CoCNet (\ie, ``bokeh''), gradient-based attacked bokeh image (\ie, ``gda''), smooth gradient-based attacked bokeh image (\ie, ``sm-gda''), background-guided gradient-based attacked bokeh image (\ie, ``bg-gda''). Here we use the projected gradient descent \cite{madry2017towards} (PGD) to implement the gradient-based attack method in the bokeh rendering. For gradient-based attacked bokeh image, smooth gradient-based attacked bokeh image and background-guided gradient-based attacked bokeh image, the maximum perturbation for each pixel (\ie, $\epsilon$) is 0.0003, 0.008 and 0.0006, respectively. The number of attack iterations is 50. For 
Compared with the natural degradation, our method significantly decreases the accuracy to less than 10\% on all four models. With a similar decrease, we further calculate the image similarity between bokeh images and attacked bokeh images. We can find that ``sm-gda'' always achieves more similar results than ``gda''. ``bg-gda'' is a little worse than them.

\textit{Transferability.}
To fully demonstrate the attack methods, we evaluate their transferability. As shown in Table~\ref{Table:attack_transferability}, the first row shows our attack methods (\ie, gda, bg-gda, sm-gda). They all attack one of the four models (\ie, ResNet50, VGG, DenseNet and MobileNetV2) and test on the other three models. The values are the test accuracy and the lower value means better transferability of the attack method. We can find that ``bg-gda'' has the best attack transferability. To summarize, the ``sm-gda'' and ``bg-gda'' have better attack performance and attack transferability than ``gda'' respectively, which shows the improvement on ``gda'' proposed by us is effective. 

\begin{figure}
\centering
\includegraphics[width=0.9\linewidth]{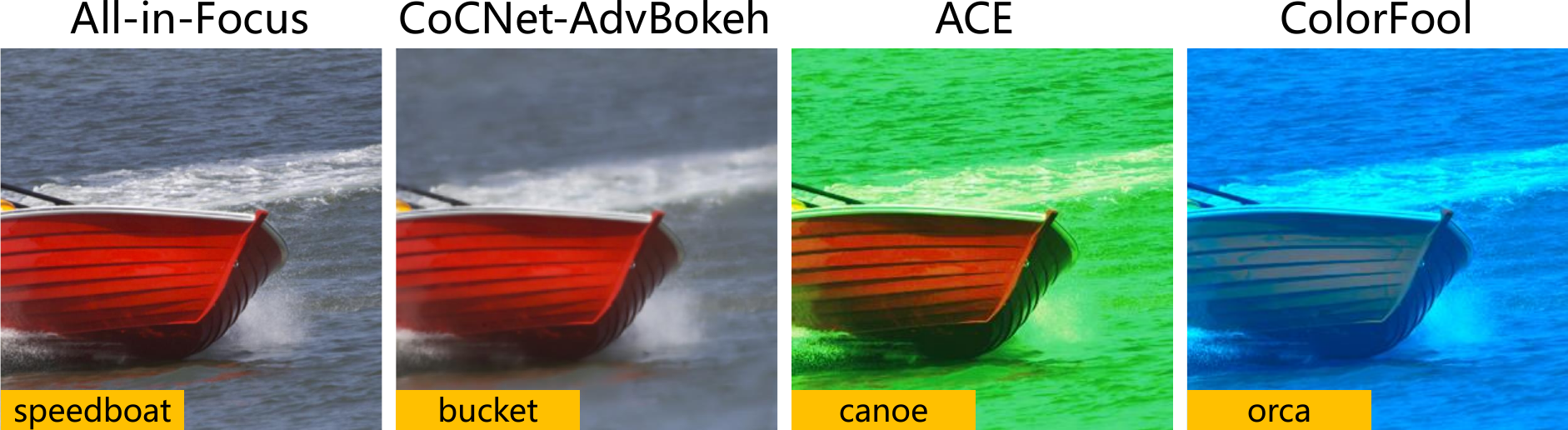}
\caption{Comparison of adversarial examples towards ResNet50 generated by our attack method with other state-of-the-art unrestricted attack methods (\ie, ColorFool \cite{shamsabadi2020colorfool} and ACE \cite{zhao2020adversarial}).}
\label{fig:compare_to_other_unrestrict}
\end{figure}

\textit{Comparion to other unrestricted attacks.}
There are some unrestricted attacks proposed to modify the semantics of the image. However, most of them lead to unnatural and absurd adversarial examples. As shown in Fig.~\ref{fig:compare_to_other_unrestrict}, we can find that the adversarial bokeh examples proposed by ``gda'' look far more natural than the other two unrestricted attacks (\ie, ColorFool and ACE) and less likely to arouse suspicion.

\begin{table}
\centering
\caption{Comparison of the defocus deblurring performance between IFAN \cite{Lee_2021_CVPR} and our finetuned IFAN (\ie, F-IFAN).}
\resizebox{0.7\linewidth}{!}{
\begin{tabular}{l|ccc}
\toprule
 & PSNR $\uparrow$ & SSIM $\uparrow$ & LPIPS $\downarrow$\tabularnewline
\midrule
IFAN & 25.36620 & 0.78885 & 0.21739\tabularnewline
F-IFAN (ours) & 25.38111 & 0.78887 & 0.21836\tabularnewline
\bottomrule
\end{tabular}
}
\label{Table:IFAN_finetune}
\end{table}
\subsection{Defocus Deblurring Task Improvement}\label{exp:down_stream_task}
Every coin has two sides. Although the adversarial bokeh examples have a bad influence on visual tasks, they can be used to improve downstream tasks. Here we take the state-of-the-art defocus deblurring method IFAN \cite{Lee_2021_CVPR} as an example. The training dataset used by them is DPDD. We use our method to generate adversarial attacked bokeh images according to the training images of DPDD. Then we collect the attacked images and original training images together as a kind of data augmentation to fine-tune the IFAN model. As shown in Table~\ref{Table:IFAN_finetune}, we compare the IFAN model with our fine-tuned model (\ie, ``F-IFAN'') on the test dataset of DPDD. We can find that the deblurred images generated by our model achieve higher similarity with the ground truth all-in-focus images than IFAN, which shows the effectiveness of our attack method in improving the defocus deblurring task.

\section{Conclusions.}\label{sec:concl}
In this paper, we propose a circle-of-confusion predictive network (CoCNet) that follows the physical priori. Based on CoCNet, we are able to research the influence of natural \& adversarial bokeh effects by revealing the vulnerability of the neural networks in visual understanding tasks. Furthermore, we demonstrate the positive usage of the attacked images as data augmentation to improve the downstream tasks. In the future, we aim to introduce GAN \cite{emami2020spa,tan2023alr,zhang2023transformer} into the framework for more realistic bokeh rendering.

\bibliographystyle{IEEEtran}
\bibliography{ref}

\end{document}